\documentclass[10pt,twocolumn,letterpaper]{article}

\usepackage{cvpr}              

\usepackage{graphicx}
\usepackage{amsmath}
\usepackage{amssymb}
\usepackage{booktabs}
\usepackage{array}
\usepackage{color}
\usepackage[dvipsnames]{xcolor}
\usepackage{etoolbox,lipsum}
\usepackage{gensymb}
\newcommand{\PreserveBackslash}[1]{\let\temp=\\#1\let\\=\temp}
\newcolumntype{C}[1]{>{\PreserveBackslash\centering}p{#1}}

\usepackage[pagebackref,breaklinks,colorlinks]{hyperref}

\definecolor{r1c}{rgb}{0.8941,0.1097,0.1020}
\definecolor{r2c}{rgb}{0.302,0.6862,0.2902}
\definecolor{r3c}{rgb}{0.2157,0.4941,0.7216}
\definecolor{darkgolden}{rgb}{0.7,0.7,0.1}
\definecolor{darkorange}{rgb}{0.9,0.5,0.1}
\definecolor{darkred}{rgb}{0.95,0.05,0.05}

\usepackage[capitalize]{cleveref}
\crefname{section}{Sec.}{Secs.}
\Crefname{section}{Section}{Sections}
\Crefname{table}{Table}{Tables}
\crefname{table}{Tab.}{Tabs.}

\def\cvprPaperID{580} 
\def\confName{CVPR}
\def\confYear{2023}

\begin{document}

\title{ShapeClipper: Scalable 3D Shape Learning from Single-View Images \\ via Geometric and CLIP-based Consistency}

\author{Zixuan Huang$^1$ \quad Varun Jampani$^{2}$ \quad Anh Thai$^{1}$ \quad Yuanzhen Li$^{2}$\\
 \quad Stefan Stojanov$^{1}$ \quad James M. Rehg$^{1}$\\
$^1$Georgia Institute of Technology, $^2$Google Research
}

\twocolumn[{%
\renewcommand\twocolumn[1][]{#1}%
\maketitle
\begin{center}
    \centering
    \captionsetup{type=figure}
    \vspace{-5mm}
    \includegraphics[width=1\textwidth]{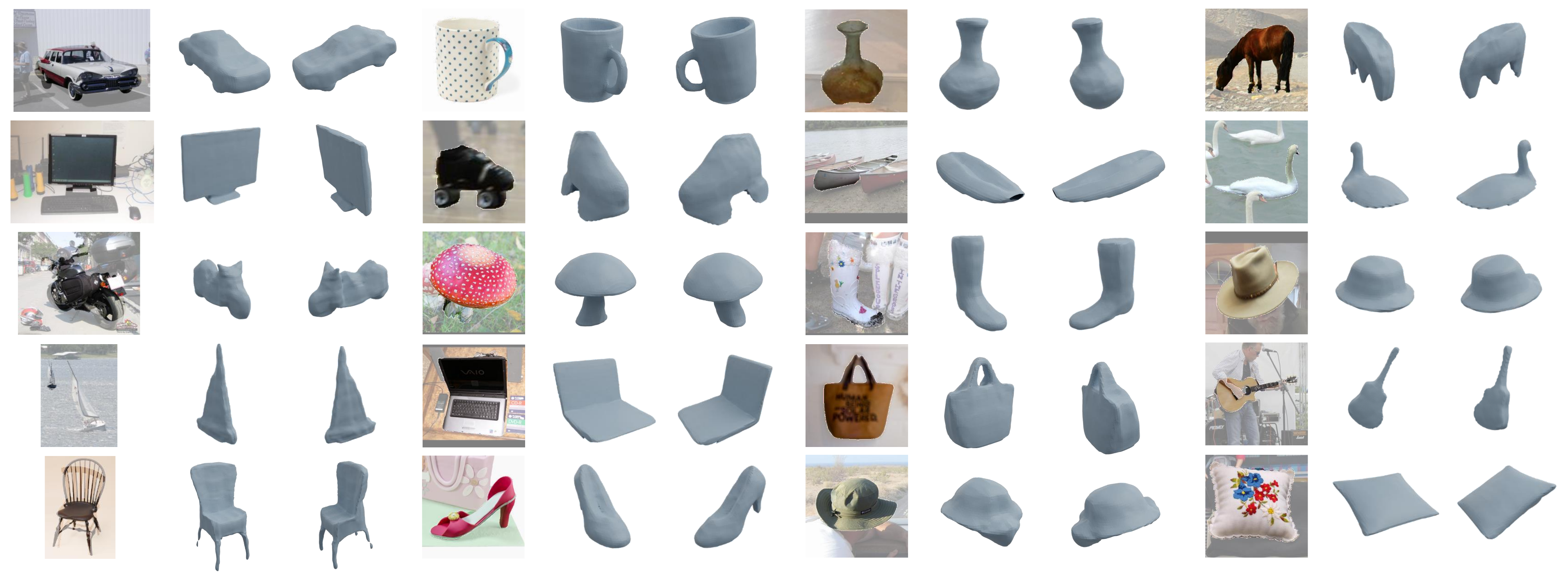}
    \captionof{figure}{We propose a method that reconstructs 3D object shape from single view real-world images. Our method learns high-quality reconstruction through \textbf{\emph{single-view supervision without known viewpoint}} and can reconstruct shapes of various objects.}
	\label{fig:result_teasor}
\end{center}%
}]


\begin{abstract}
We present ShapeClipper, a novel method that reconstructs 3D object shapes from real-world single-view RGB images.
Instead of relying on laborious 3D, multi-view or camera pose annotation, ShapeClipper learns shape reconstruction from a set of single-view segmented images. 
The key idea is to facilitate shape learning via CLIP-based shape consistency, where we encourage objects with similar CLIP encodings to share similar shapes. 
We also leverage off-the-shelf normals as an additional geometric constraint so the model can learn better bottom-up reasoning of detailed surface geometry.
These two novel consistency constraints, when used to regularize our model, improve its ability to learn  both global shape structure and local geometric details.
We evaluate our method over three challenging real-world datasets, Pix3D, Pascal3D+, and OpenImages, where we achieve superior performance over state-of-the-art methods.\footnote{Project website at: \url{https://zixuanh.com/projects/shapeclipper.html}}
\end{abstract}


\section{Introduction}
\label{sec:intro}


How can we learn 3D shape reconstruction from real-world images in a scalable way? Recent works achieved impressive results via learning-based approaches either with 3D~\cite{choy20163d,Gkioxari_2019_ICCV,groueix2018,mescheder2019occupancy,thai20213d,saito2019pifu,xu2019disn,wu2017marrnet,wang2018pixel2mesh,wen2019pixel2mesh++,xie2019pix2vox,yao2020front2back} or multi-view supervision~\cite{yan2016perspective,tulsiani2017multi,yang2018learning,lin2018learning,insafutdinov2018unsupervised,kato2018neural,liu2019soft,niemeyer2020differentiable}. However, such supervised techniques cannot be easily applied to real-world scenarios, because it is expensive to obtain 3D or multi-view supervision at a large scale. To address this limitation, recent works 
relax the requirement for 3D or multi-view supervision~\cite{kanazawa2018learning,goel2020shape,li2020self,gadelha20173d,henzler2019escaping,ye2021shelf,navaneet2020image,kato2019learning,wu2020unsupervised,lin2020sdf,huang2022planes,alwala2022pre,monnier2022unicorn,simoni2021multi,zhang2022monocular}. These works only require single-view self-supervision, with some additionally using expensive viewpoint annotations~\cite{kanazawa2018learning,kato2019learning,lin2020sdf,simoni2021multi,zhang2022monocular}. Despite this significant progress, most methods still suffer from two major limitations: 1) Incorrect top-down reasoning, where the model only explains the input view but does not accurately reconstruct the full 3D object shape;
2) Failed bottom-up reasoning, where the model cannot capture low-level geometric details such as concavities.
How can we address these limitations while also remaining scalable to a wide range of object types?


To improve top-down reasoning, our inspiration comes from the recent success of large-scale image-text modeling.
The most successful image-text models such as CLIP~\cite{radford2021learning} are trained on a vast corpus of captioned images and are able to extract fine-grained semantic features that correlate well with the language descriptions.
CLIP further demonstrates a great generalization ability to images across various domains. \textit{Can we leverage such a powerful and generalizable model to learn 3D reconstruction in a real-world scenario?}

\begin{figure}[t]
\centering
	\includegraphics[width=1\linewidth]{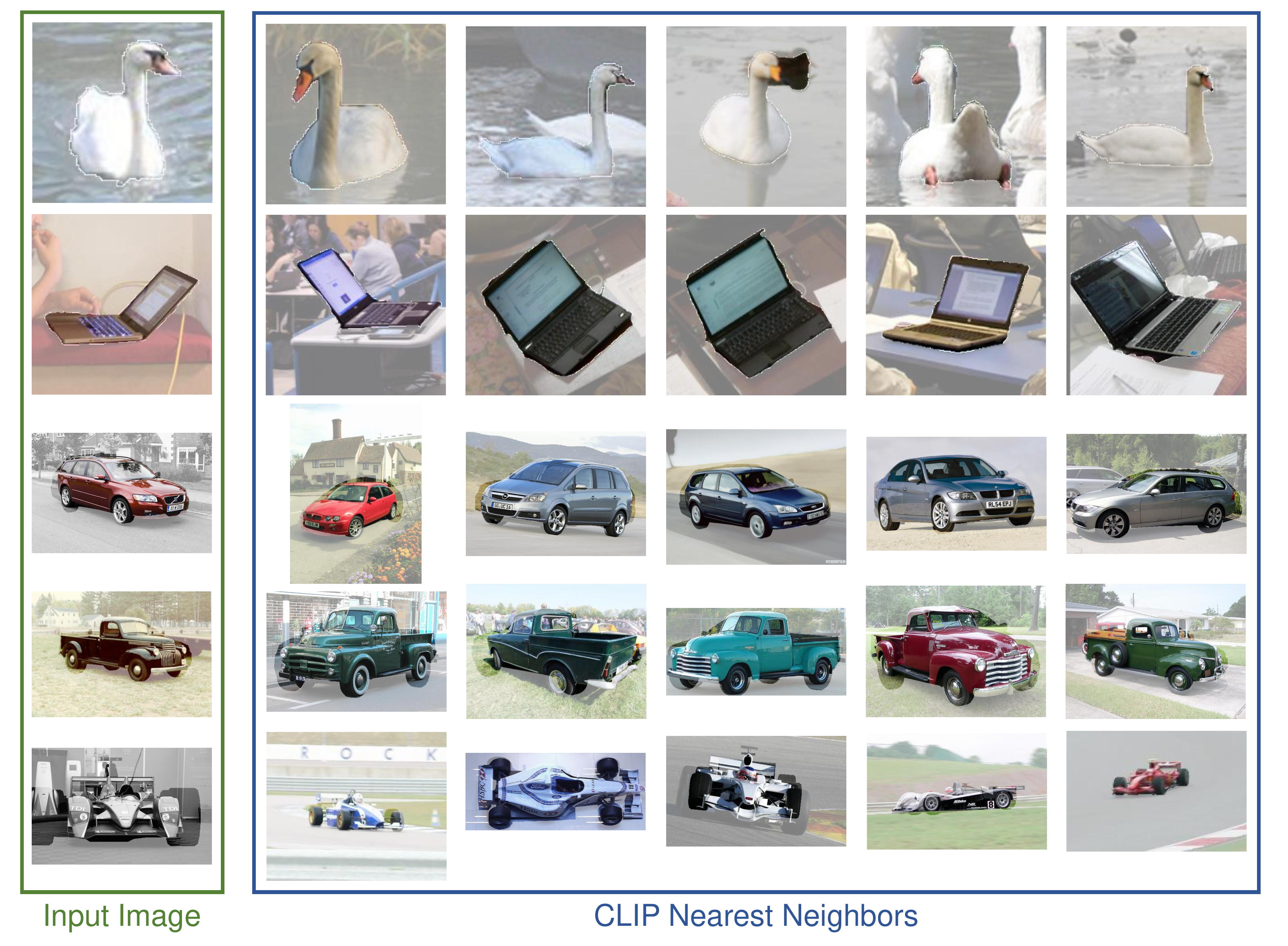}
	\caption{\textbf{CLIP-based semantic neighbors.} Samples that have similar CLIP encodings often have similar shapes. Note the viewpoint variability in the neighbors.}
	\label{fig:nn_teasor}
	\vspace{-16pt}
\end{figure}
We observe that natural language descriptions of images often contain geometry-related information (\eg a \textit{round} speaker, a \textit{long} bench) and many nouns by themselves have characteristic shape properties (\eg “desks” usually have four legs, and “benches” normally include a flat surface). Motivated by this intrinsic connection between object shapes and language-based semantics, we examine the latent space of CLIP's visual encoder. In our study, we find (via k-NN queries) that objects with similar CLIP embeddings usually share similar shapes (see~\cref{fig:nn_teasor} for an example). 
Another key characteristic we identify with CLIP embeddings is that they have some robustness to viewpoint changes, meaning that changes in viewpoint generally do not produce drastic changes in CLIP embeddings.

\begin{figure}[t]
\centering
	\includegraphics[width=1\linewidth]{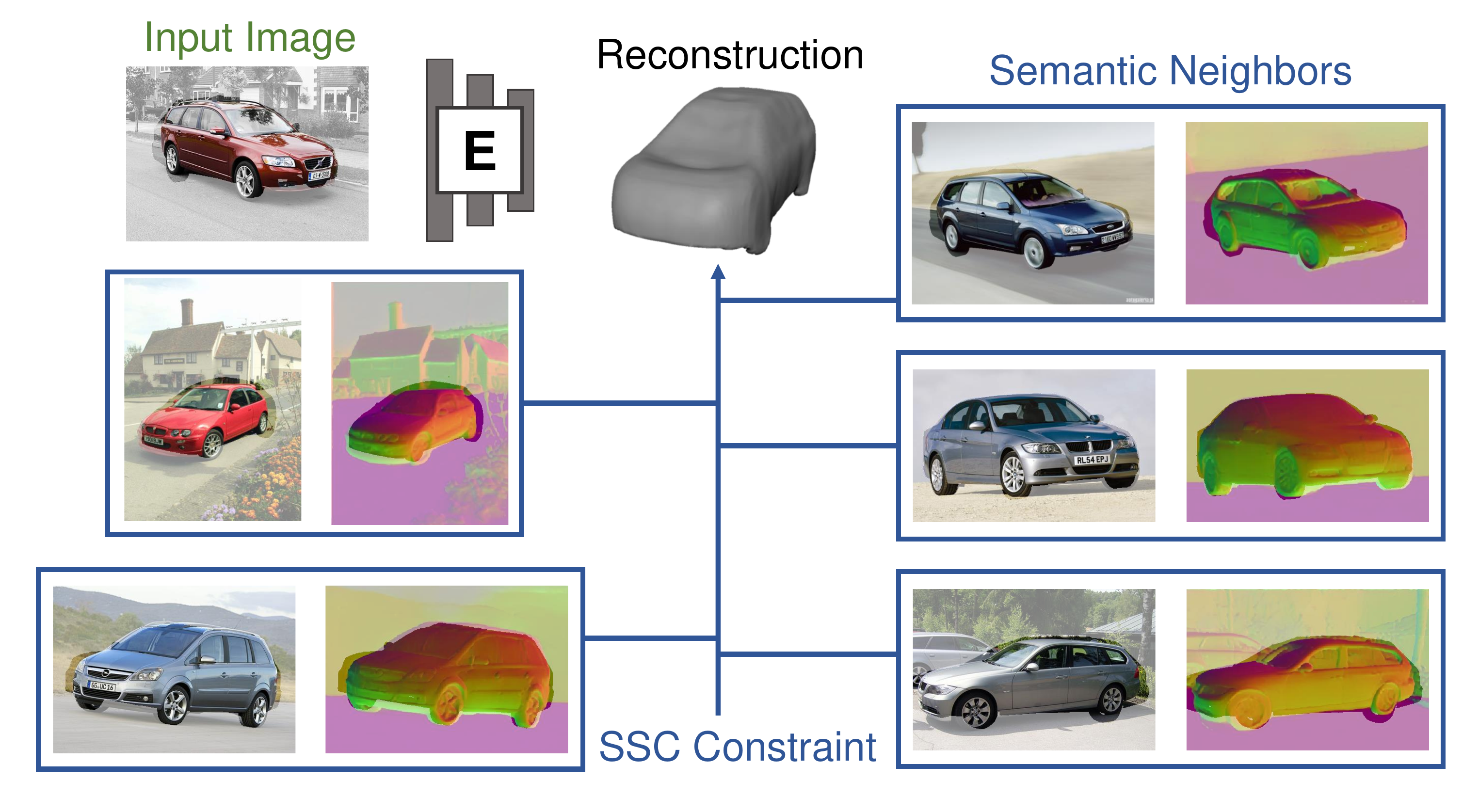}
	\caption{\textbf{Semantic-based Shape Consistency (SSC) Constraint.} We find the semantic neighbors of the input image across the training set and use these neighbors to regularize the shape learning.}
	\label{fig:clip_teasor}
	\vspace{-16pt}
\end{figure}

Inspired by these findings, we propose to learn shapes using a semantic-based shape consistency (SSC) constraint using CLIP. Specifically, we use CLIP's semantic encodings as guidance to form pseudo multi-view image sets. For each image in the training set, we extract its CLIP embedding and find images with the most similar semantics across the training set. We then leverage these retrieved images as additional supervision to the input view, as illustrated in~\cref{fig:clip_teasor}. This approach greatly benefits global shape understanding, because each predicted shape is required to simultaneously explain a set of images instead of only explaining the single input image.

On the other hand, we address the limitation of poor bottom-up geometric reasoning by constraining the surface normals of the predicted shapes. Common failure cases include noisy surface reconstruction and failed concavity modeling, which are extremely hard to learn even with multi-view supervision. Inspired by the recent success of large-scale 2.5D estimation that generalizes to various scenes~\cite{eftekhar2021omnidata,ranftl2020towards,ranftl2021vision}, we propose to use off-the-shelf surface normals as additional geometric supervision for our task. However, unlike scenes, off-the-shelf normals for object-centric images are much noisier due to occlusion, truncation, and domain gaps. To mitigate this issue, we introduce a noise-tolerant optimization process via outlier dropout, which stabilizes the training and improves the overall reconstruction performance.

Overall, our contributions are threefold:
\begin{itemize}
\setlength\itemsep{-0.2em}
    \item We propose a novel CLIP-based shape consistency regularization that greatly facilitates the top-down understanding of object shapes.
    \item We successfully leverage off-the-shelf geometry cues to improve single-view object shape reconstruction for the first time and handle noise effectively.
    \item We perform extensive experiments across 3 different real-world datasets and demonstrate state-of-the-art performance.
\end{itemize}

\section{Related Work}
\label{sec:related}

There has been an emerging interest in 3D object shape reconstruction from images via learning-based approaches. Our work focuses on learning single-view shape reconstruction with limited supervision on real-world images, where the training set only contains a single view per object instance. We briefly survey the relevant literature on single image shape reconstruction using both fully-supervised and weakly-supervised approaches. 

\noindent \textbf{Single-View Supervision.} 
Most closely related to this paper are works that learn 3D shape reconstruction through supervision from single-view images ~\cite{kanazawa2018learning,goel2020shape,li2020self,henzler2019escaping,henderson2019learning,ye2021shelf,navaneet2020image,kato2019learning,wu2020unsupervised,lin2020sdf,zhang2022monocular,huang2022planes,alwala2022pre,monnier2022unicorn}. 
These works can be organized as in \cref{table:method-comparison} and largely differ in their choice of 1) shape representation (e.g. implicit SDF vs explicit mesh); 2) known vs. unknown viewpoint supervision; 3) large-scale evaluation on various real-world objects. We are one of the earliest works to demonstrate the feasibility of single-view learning of an implicit SDF representation upon diverse real-world images under unknown viewpoints. 

\begin{table*}
    \centering
    \caption{\textbf{Single-view supervised methods for object shape reconstruction.} M: mesh, V: voxel, P: pointcloud, D: depth, O: occupancy function, S: signed distance function, Diverse R-Res.: real-world results on diverse categories.}
    \vspace{-7pt}
    \small
    \begin{tabular}{|c|c|c|c|c|c|c|c|c|c|c|c|c|c|c|c|c|}  \hline
    Model & \cite{kanazawa2018learning} & \cite{kato2019learning} & \cite{lin2020sdf} & \cite{simoni2021multi} & \cite{zhang2022monocular} & \cite{goel2020shape} & \cite{li2020self} & \cite{henzler2019escaping} & \cite{henderson2019learning} & \cite{navaneet2020image} & \cite{wu2020unsupervised} & \cite{monnier2022unicorn} & \cite{huang2022planes} & \cite{ye2021shelf} & \cite{alwala2022pre}  & Ours \\ \hline 
    3D Rep. & M & M & S & M & M & M & M & V & M & P & D & M & S & M & O & S \\
    Viewpoint Free & - & - & - & - & - & \checkmark & \checkmark & \checkmark & \checkmark & \checkmark & \checkmark & \checkmark & \checkmark & \checkmark & \checkmark & \checkmark \\
    Diverse R-Res. & - & - & - & - & - & - & - & - & - & - & - & - & - & \checkmark & \checkmark & \checkmark \\
    \hline
    \end{tabular}
    \label{table:method-comparison}
    \vspace{-12pt}
\end{table*}

Within this body of work, SSMP~\cite{ye2021shelf}, Cat3D~\cite{huang2022planes}, and SS3D~\cite{alwala2022pre} are the most closely related ones given their large-scale evaluation, which we describe in details below.

SSMP~\cite{ye2021shelf} is the earliest work that shows success of shape learning via only single-view supervision on large-scale real-world data. A key property of this method is adversarial regularization during training, which can make training unstable. Thus it is hard for SSMP to learn reconstruction on categories with complex shapes or textures. In contrast, our method leverages the SSC and geometric constraints which are more stable and result in superior performance over SSMP across various objects. 

Similar to our method, Cat3D~\cite{huang2022planes} explores semantic regularization for implicit shape learning. In contrast, their semantic regularization is based on category labels, which fails on categories with significant intra-category shape variations. Moreover, Cat3D relies on unstable adversarial regularization which has similar drawbacks as SSMP~\cite{ye2021shelf} and is only successful on a few real-world categories. 

SS3D~\cite{alwala2022pre} proposes a 3-step learning pipeline for scalable learning of shapes, which includes synthetic data (e.g. ShapeNet~\cite{chang2015shapenet}) pretraining. This step plays a crucial role as it provides the necessary initialization for the camera multiplex optimization. Unlike SS3D, synthetic pretraining is not a hard constraint for our method---we demonstrate success on Pix3D~\cite{pix3d} without any synthetic pretraining. On the other hand, SS3D does not explore the usage of semantic and geometric consistency. As a result, our model captures both global structures and local surfaces more accurately than SS3D and outperforms SS3D quantitatively. 

\noindent \textbf{Shape Supervision.} 
Instead of using image supervision, many prior works use explicit 3D geometric supervision and achieve great reconstruction results~\cite{choy20163d,Gkioxari_2019_ICCV,groueix2018,mescheder2019occupancy,thai20213d,saito2019pifu,xu2019disn,wu2017marrnet,wang2018pixel2mesh,wen2019pixel2mesh++,xie2019pix2vox,yao2020front2back}. Nevertheless, the assumption of 3D supervision is not yet practical on a large scale. To make the learning more scalable, subsequent works leverage multi-view images as supervision and employ differentiable rendering as the core technique. Specifically, differentiable rendering allows images and masks to be rendered from 3D assets differentiably and thus the multi-view reprojection loss can effectively carve the reconstructed shape. These methods can be classified based on their representation of shape, including voxels~\cite{yan2016perspective,tulsiani2017multi,yang2018learning}, pointclouds~\cite{lin2018learning,insafutdinov2018unsupervised}, meshes~\cite{kato2018neural,liu2019soft} and implicit representations~\cite{niemeyer2020differentiable}. Compared to these works, a major benefit of our method is scalability, as our model can be trained using single-view real-world images.


\section{Method}
\label{sec:method}

In this section, we first present an overview of our model in~\cref{subsection:overview}, and then introduce our proposed SSC and geometric constraints in~\cref{subsection:semantic} and~\cref{subsection:geometric}. Finally, we present implementation details in~\cref{subsection:implementation}.

\subsection{Overview}
\label{subsection:overview}

\begin{figure}[t]
\centering
	\includegraphics[width=0.95\linewidth]{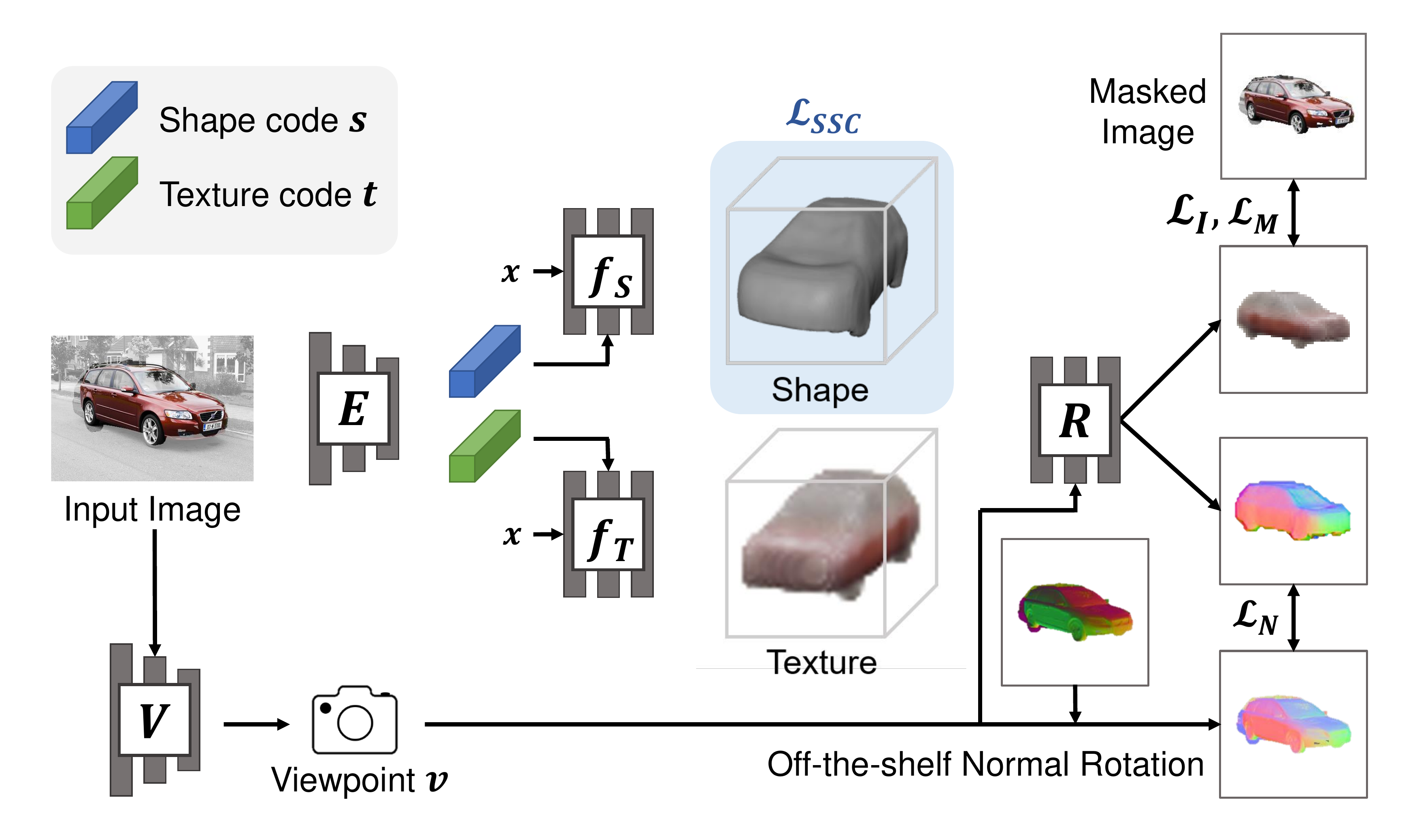}
	\vspace{-10pt}
	\caption{\textbf{Network overview.} Given the input image, the encoder $\boldsymbol{E}$ infers the shape latent code $\boldsymbol{s}$ and texture latent code $\boldsymbol{t}$. By conditioning these two codes upon shape MLP $\boldsymbol{f_S}$ and texture MLP $\boldsymbol{f_T}$, we obtain the shape and texture reconstruction of the input object. On the other hand, the viewpoint estimator $\boldsymbol{V}$ estimates the input viewpoint $\boldsymbol{v}$. The differentiable volume renderer $\boldsymbol{R}$ then renders shape and texture fields under the estimated viewpoint, so that we can compute the reconstruction loss $\mathcal{L}_I$ and $\mathcal{L}_M$. We further leverage our SSC and geometric constraints, $\mathcal{L}_{SSC}$ and $\mathcal{L}_N$, to effectively harness the shape learning.}
	\label{fig:arch}
	\vspace{-17.5pt}
\end{figure}

Given a collection of $n$ images segmented with foreground masks $\{I_i\in\mathbb{R}^{h \times w \times 3}, M_i\in\mathbb{R}^{h \times w \times 1}\}_{i=1}^n$, we aim to learn a single-view 3D reconstruction model without relying on 3D, viewpoint, or multi-view supervision of these images.
The shape representation of our model is an implicit SDF function, represented by a multi-layer perceptron (MLP) conditioned on the input image. Specifically, our model consists of four submodules (see~\cref{fig:arch} for an overview) as described below.

\noindent \textbf{Image encoder.} 
The image encoder $\boldsymbol{E}$ takes a segmented image $I \in \mathbb{R}^{h \times w \times 3}$ as input and infers the shape latent code $\boldsymbol{s} \in \mathbb{R}^{l}$ and the texture latent code $\boldsymbol{t} \in \mathbb{R}^{l}$. These two codes encode the necessary information to reconstruct the shape and texture field respectively.

\noindent \textbf{Shape and texture reconstructor.} 
Our model represents shape and texture reconstruction with two MLPs, $\boldsymbol{f_S}: \mathbb{R}^3 \rightarrow \mathbb{R}$ and $\boldsymbol{f_T}: \mathbb{R}^3 \rightarrow \mathbb{R}^3$, which predict signed distance function (SDF) and RGB values of queried 3D coordinates respectively. The MLPs are conditioned on the latent codes, with a similar design to VolSDF~\cite{yariv2021volume}. Specifically, the shape MLP $\boldsymbol{f_S}$ is conditioned on $\boldsymbol{s}$ and the texture MLP $\boldsymbol{f_T}$ is conditioned on $\boldsymbol{t}$.

\noindent \textbf{Viewpoint estimator.} 
The viewpoint estimator $\boldsymbol{V}$ estimates the viewpoint of the input image with respect to the shape reconstruction. Following~\cite{beyer2015biternion,huang2022planes}, we represent the viewpoint with the trigonometric functions of Euler angles, i.e. $\boldsymbol{v} = [\cos \boldsymbol{\gamma}, \sin \boldsymbol{\gamma}]$ where $\boldsymbol{\gamma}$ denotes the three Euler angles. 

\noindent \textbf{Differentiable renderer.} 
We use a volume renderer $\boldsymbol{R}$ to render the reconstructed SDF and texture fields following VolSDF~\cite{yariv2021volume}. In the renderer, the SDF field is first converted into densities, and then the densities are used together with the texture field to render the RGB and mask in an accumulative way (via ray-marching). We refer the readers to VolSDF~\cite{yariv2021volume} for more details, with the exception that we use uniform sampling instead of error-bound based sampling. Formally, we denote the renderer as a functional, $\boldsymbol{R}(\boldsymbol{f_S}, \boldsymbol{f_T}, \boldsymbol{v})$, which maps the implicit functions and the viewpoint into image $\hat{I}$, mask $\hat{M}$ and surface normal $\hat{N}$. 

\noindent \textbf{Reprojection loss.} 
One of the major learning signals of our model comes from the reprojection loss that compares input images with reconstructed images. This can be achieved via the differentiable renderer. Our model first infers shape, texture, and viewpoint of the object from the input image. The renderer can then render them into an image reconstruction, which will be matched to the input. 

Specifically, we can denote the RGB and the mask reprojection loss for each image as follows:
\begin{equation}
    \mathcal{L}_{I} = \|I-\hat{I}\|^2_2,\ \mathcal{L}_{M} = 1 - IOU(M, \hat{M}),
    \label{eq:image_mask}
\end{equation}
\vspace{-12pt}
\begin{equation}
    IOU(M, \hat{M}) = \frac{\sum_{p} M^p \cdot \hat{M^p}}{\sum_{p} M^p + \hat{M^p} - M \cdot \hat{M^p}}.
    \label{eq:iou}
\end{equation}
Here $M^p$ denotes the mask value at pixel $p$.

\noindent \textbf{Facilitating shape learning.} 
When we do not have direct viewpoint or shape supervision, simply minimizing the reprojection loss almost always leads to degeneration. There are two major issues: 1) incorrect top-down reasoning, where shapes can only explain the input view; 2) wrong bottom-up reasoning, examples include the inability to infer concavity or noisy surface reconstruction. To mitigate these issues, we propose the semantic and geometric consistency constraints that effectively facilitate the shape learning.  

\subsection{Semantic Constraint}
\label{subsection:semantic}

\noindent \textbf{Preliminary findings about CLIP.} 
To leverage CLIP for shape regularization, our main hypothesis is that objects with similar CLIP encodings share similar shapes. To verify this hypothesis, we perform a study using the large-scale fine-grained CompCars~\cite{yang2015large} dataset. This dataset contains more than 136K images of 163 car makes with 1716 car models. We perform CLIP inference on this dataset and compute 5-nearest neighbor for each sample based on the CLIP embeddings. By iterating over each neighbor of all query images, we calculate the percentage of neighbors that match their query images' model (same car model usually shares quite similar shapes). In our experiment, CLIP is able to find the exact same car models for 51.2\% of all the neighbors (on average 2.6 out of 5 neighbors belong to the query images' model). We believe this is a promising finding given 1) the large number of images and models in CompCars and 2) the fact that different models can still have similar shapes, so the percentage of shape matches can be higher than exact model matches. As a comparison, the percentage of model matches for ImageNet-pretrained ViT~\cite{dosovitskiy2020image} is only 27.8\%. This study verifies our hypothesis and enables us to design our Semantic-based Shape Consistency (SSC) constraint based on CLIP.

\noindent \textbf{Semantic-based Shape Consistency.} 
The key idea of SSC is to pull instances with similar CLIP embeddings together, so that a single shape reconstruction can receive supervision from all these instances. 
In our experiments, we find CLIP encodings have some robustness to viewpoint change (see supplement for more details). Therefore it enables us to find additional pseudo views for many objects, which significantly facilitate the learning of better top-down reasoning.

We first form the per-instance clusters by performing K-nearest neighbors with CLIP encodings. Formally, given our training set $\{I_i\}$, we extract the CLIP encoding for each image, denoted as $\{\boldsymbol{c}_i\}$. We calculate the cosine similarity of all pairs of encodings. With such similarity measurement, we can query the K-nearest neighbors for any specific input encoding $\{\boldsymbol{c}_i\}$ and identify images and masks of these neighbors. 

\begin{figure}[t]
\centering
	\includegraphics[width=0.95\linewidth]{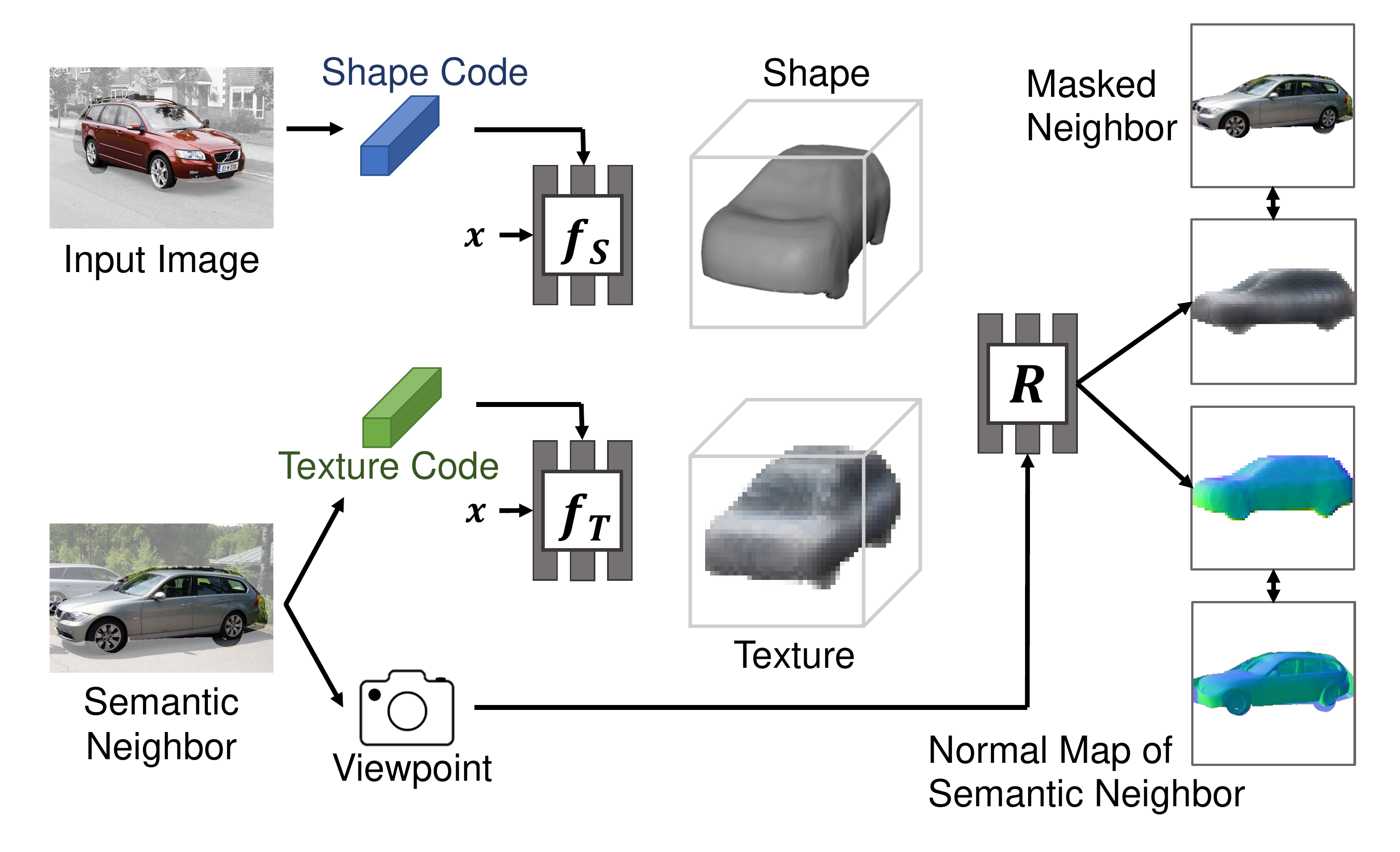}
	\vspace{-10pt}
	\caption{\textbf{Semantic-based Shape Consistency (SSC) constraint.} We improve shape learning via the SSC constraint. The shape reconstructed for the input object has to explain its CLIP semantic neighbor as well.}
	\label{fig:ssc}
	\vspace{-17pt}
\end{figure}
We then use these semantic neighbors to supervise the shape reconstruction as in~\cref{fig:ssc}. The high-level idea is that 1) the shape reconstructed for the input should explain neighbors' masks and normals, and 2) when combining input's shape with neighbor's texture, we should be able to render the neighbor image. 

Formally, for input $I$, we denote its shape latent code and shape MLP as $\boldsymbol{s}$ and $\boldsymbol{f_S}$. Meanwhile, we sample an image and its mask $I_{k}$ and $M_{k}$ from the semantic neighbor set $\{I_{k}, M_{k}\}^K_{k=1}$. 
The encoder $\boldsymbol{E}$ then predicts the latent texture code $\boldsymbol{t}_{k}$ for $I_{k}$, which is used to generate its texture function $\boldsymbol{f_T}_k$. We also obtain the viewpoint prediction $\boldsymbol{v}_{k}$ through the viewpoint estimator $\boldsymbol{V}$. We can then combine the input shape $\boldsymbol{f_S}$ with the neighbor texture $\boldsymbol{f_T}_k$ and viewpoint $\boldsymbol{v}_{k}$, and render them into an image $\hat{I}'$, a mask $\hat{M}'$ and a normal map $\hat{N}'$. Namely, $(\hat{I}', \hat{M}', \hat{N}') = \boldsymbol{R}(\boldsymbol{f_S}, \boldsymbol{f_T}_k, \boldsymbol{v}_{k})$. By replacing input maps in~\cref{eq:image_mask} with semantic neighbors, we can obtain the SSC losses for this sample in the same form:
\vspace{-5pt}
\begin{equation}
    \mathcal{L}_{SSC_I} = \|I_k-\hat{I}'\|^2_2,\  \mathcal{L}_{SSC_M} = 1 - IOU({M_k}, \hat{M'}).
    \label{eq:ssc_im}
\end{equation}


\subsection{Geometric Constraint}
\label{subsection:geometric}
We further propose to facilitate shape learning via geometric constraints to encourage the model to learn better low-level geometric reasoning. The idea here is to estimate the surface normal of our implicit shape, and make it consistent with the surface normal prediction from off-the-shelf models. The off-the-shelf normal estimator we use is Omnidata~\cite{eftekhar2021omnidata}, which is a state-of-the-art normal estimator.
Recent work has also proven its effectiveness for multi-view scene reconstruction~\cite{yu2022monosdf}. 

Formally, we denote our surface normal estimation as $\hat{N} \in \mathbb{R}^{h \times w \times 3}$ and the off-the-shelf unit normal as $N \in \mathbb{R}^{h \times w \times 3}$. The estimated normal is calculated as the normalized gradient of the density and aggregated via volume rendering similar to MonoSDF~\cite{yu2022monosdf}. Unlike the setup in MonoSDF, our normal estimation lies in the object-centric canonical frame instead of the view-centric frame. Therefore, we use our estimated viewpoint to rotate the off-the-shelf normal $N$ to be in the same canonical frame as $\hat{N}$. 
In addition to aligning the coordinate frames, this approach enables the viewpoint estimator to receive additional training signals from local geometry alignment, which is a significant benefit that naive approaches like the closed-form rotation alignment cannot provide. After the rotation, we can then match the normals following~\cite{eftekhar2021omnidata}:

\begin{equation}
    \mathcal{L}_{N} = \beta \cdot \|RN-\hat{N}\|_1 - cos(RN, \hat{N}),
\end{equation}
where $R$ refers to the rotation matrix derived from the estimated viewpoint and $cos$ denotes the cosine similarity. We set $\beta = 5$ across all the experiments.

This geometric loss is calculated at the pixel level and averaged over a minibatch. However, unlike scene reconstruction~\cite{yu2022monosdf}, off-the-shelf normals can be noisy for object-centric images due to inaccurate masks and domain gaps. As a result, naively using off-the-shelf normals results in training instability. Inspired by online hard example mining~\cite{shrivastava2016training}, we propose to dropout off-the-shelf normals that are likely to be outliers via batchwise ranking. Specifically, we sort the normal loss $\mathcal{L}_{N}$ within the current minibatch and exclude a fixed percentage of high-loss pixels from the final loss aggregation. We find that this strategy stabilizes the training and improves the reconstruction quality overall.


Finally, we can combine the geometric constraint with the semantic constraint by having a SSC normal loss, $\mathcal{L}_{SSC_N}$. This can be calculated similarly to $\mathcal{L}_{N}$, the only difference is that we replace the input off-the-shelf normals and rotations with the semantic neighbor's as in~\cref{eq:ssc_im}.

\subsection{Implementation Details}
\label{subsection:implementation}
\noindent \textbf{Architecture.}
The image encoder we use is a ResNet34~\cite{he2016deep}, which projects the input image into two 64-d latent vectors representing shape and texture. We use lightweight MLPs to represent the SDF and texture fields, where the shape MLP has 5 hidden layers of 64 neurons and the texture MLP has 3 hidden layers of 64 neurons. The 3D coordinates are positionally encoded~\cite{martin2020nerf} before fed into the MLPs. The conditioning of the MLPs is achieved via concatenation, and the shape latent code is additionally skip-connected to the first and the second hidden layers of the shape MLP. Following VolSDF, we condition the texture MLP with the shape MLP's last-layer feature as well. The differentiable renderer we use renders the volumes by uniformly sampling 64 points along each ray.

\noindent \textbf{Loss function.} 
Our overall loss function is a summation of the reconstruction loss and the SSC losses (with our geometric constraint included):
\begin{equation}
    \mathcal{L}_{recon} = \mathcal{L}_{I} + \lambda_1 \mathcal{L}_{M} + \lambda_2 \mathcal{L}_{N},
\end{equation}
\begin{equation}
    \mathcal{L}_{SSC} = \mathcal{L}_{SSC_I} + \lambda_1 \mathcal{L}_{SSC_M} + \lambda_2 \mathcal{L}_{SSC_N},
\end{equation}
\begin{equation}
    \mathcal{L} = \mathcal{L}_{recon} + \mathcal{L}_{SSC}.
\end{equation}

We set $\lambda_1 = 0.5$ and $\lambda_2 = 0.01$ across all datasets.

\noindent \textbf{Training.}
We use the Adam~\cite{kingma2014adam} optimizer with a learning rate of 0.0001 and a batch size of 12. We did not use weight decay or learning rate scheduling. Instead of using all pixels at once, we sample 512 rays to perform ray-marching for each image. Our model is trained on a single NVIDIA GTX TITAN Xp for 200 to 400 epochs depending on the dataset size, which usually takes 1 to 3 days to train. Following SS3D~\cite{alwala2022pre}, we initialize the model by pretraining on ShapeNet for our experiments on Pascal3D+ and OpenImages, where we compare to SS3D. We use the commonly used symmetry constraint~\cite{mustikovela2020self,ye2021shelf} and regularize the azimuth with a uniform prior. We further regularize the SDF field with the eikonal loss so the gradient norm is close to 1. For Pix3D we did not use synthetic pretraining, and instead we pretrain the shape to a sphere for a better initialization similar to Cat3D~\cite{huang2022planes}. 

\section{Experiments}
\label{sec:experiments}

We present the findings of applying our method across three real-world datasets in this section, including state-of-the-art comparison and detailed ablations. We first introduce the datasets we use and then describe the evaluation metrics as well as the baselines. Finally, we show detailed experiment results on each dataset.

\subsection{Datasets} 
We evaluate our method across three real-world datasets, including Pix3D~\cite{pix3d}, Pascal3D+\cite{xiang2014beyond} and OpenImages~\cite{kuznetsova2020open}.

\noindent \textbf{Pix3D.} 
Pix3D is a real-world 3D object dataset where each image is annotated with a corresponding object mask, a CAD model, and the input viewpoint. This 3D information is obtained via manual alignment between shapes and images. We use the chair category, which is the dominant category of this dataset. We follow the 70/10/20 split of~\cite{huang2022planes}, resulting in 2007, 303, 584 images for training/validating/testing respectively. 

\noindent \textbf{Pascal3D+.} 
Pascal3D+ is a real-world 3D object dataset obtained similarly to Pix3D. Compared to Pix3D, this dataset is more challenging because 1) it includes 12 diverse categories with a high-variance viewpoint distribution, and 2) object masks are quite noisy and some objects are occluded. We use all categories in the ImageNet~\cite{deng2009imagenet} subset of this dataset, with a similar split to~\cite{lin2020sdf,huang2022planes}, resulting in 11317, 11421 images for training and testing respectively. 

\noindent \textbf{OpenImages.} 
Unlike Pix3D and Pascal3D+, OpenImages do not contain any 3D annotation, which prevents us from evaluating methods quantitatively. We use 20 diverse categories (more in supplement) to train our model and leverage the occlusion scores from~\cite{ye2021shelf} to filter out highly occluded images. Each category includes 1000 to 3000 images, which are split into 90/10 for training and testing. 

\begin{figure}[t]
\centering
	\includegraphics[width=1\linewidth]{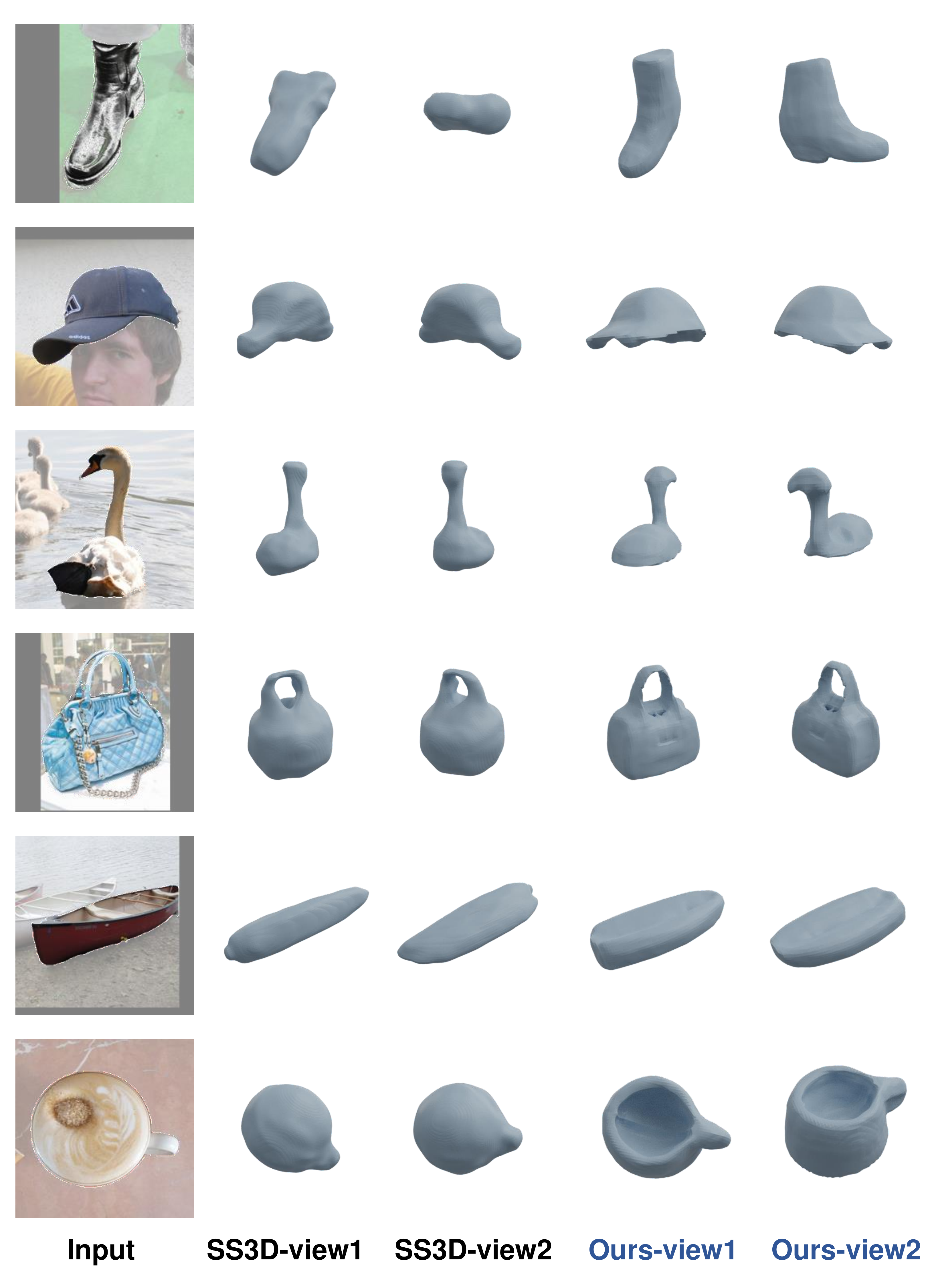}
	\caption{\textbf{Qualitative comparison on OpenImages.}  Our method learns both better global 3D structure and shape details on various categories.}
	\label{fig:openimages} 
	\vspace{-18pt}
\end{figure}

\subsection{Evaluation} 
To evaluate the implicit shapes, we sample SDF values with a $100^3$ spatial grid and extract the 0-isosurface via Marching Cubes~\cite{lorensen1987marching}. We further align the coordinate frames between predicted meshes and ground truth (GT) meshes, by transforming all meshes into the view-centric coordinate frame. We also align the scales of these meshes using the size of their projections on the image plane. After these steps, we can sample points from mesh surfaces and calculate Chamfer Distance and F-score as our quantitative metrics following~\cite{lin2020sdf,groueix2018,tatarchenko2019single,thai20213d,huang2022planes}. 

\noindent \textbf{Chamfer Distance.} 
Following~\cite{huang2022planes}, Chamfer Distance (CD) is defined as an average of accuracy and completeness. Given two pointclouds $S_1$ and $S_2$, CD can be written as:

\begin{equation}
    \small
    d(S_1, S_2) = \frac{1}{2|S_1|}\sum_{x \in S_1} \min_{y \in S_2} \|x-y\|_2 + \frac{1}{2|S_2|}\sum_{y \in S_2} \min_{x \in S_1} \|x-y\|_2
\end{equation}

\noindent \textbf{F-score.} 
F-score (FS@$d$) is a joint measurement of accuracy and completeness with a given threshold $d$. Specifically, precision@$d$ is the percentage of predicted points that have at least one GT neighbor within distance $d$. Similarly, recall@$d$ is the percentage of ground truth points that have at least one neighboring predicted points within distance $d$. FS@$d$ is then calculated as the harmonic mean of precision@$d$ and recall@$d$. It can be intuitively understood as the percentage of surface that are correctly reconstructed. 

\begin{table}
\centering
\caption{\textbf{Quantitative results on Pix3D.} Our method performs favorably to baselines and other SOTA methods. }
\begin{tabular}{lccc|c}
\hline
Methods                     & FS@1$\uparrow$        & FS@5$\uparrow$    & FS@10$\uparrow$       & CD$\downarrow$    \\ \hline
w/o $\mathcal{L}_{SSC}$     & 0.0958                & 0.4309            & 0.7093                & 0.749             \\ 
w/o normal                  & 0.0815                & 0.3913            & 0.6982                & 0.766             \\ 
w/o noise-tol               & 0.1277                & 0.5319            & 0.7861                & 0.640             \\ 
Ours                        & \textbf{0.1317}       & \textbf{0.5473}   & \textbf{0.8002}       & \textbf{0.618}    \\ \hline
Cat3D~\cite{huang2022planes}& 0.0960                & 0.4410            & 0.7262                & 0.679             \\
SSMP~\cite{ye2021shelf}     & 0.0948                & 0.4261            & 0.7168                & 0.707             \\ \hline
\end{tabular}
\label{table:pix3d}
\vspace{-15pt}
\end{table}

\subsection{Baselines} 
We consider three different baselines in this work, including SSMP~\cite{ye2021shelf}, Cat3D~\cite{huang2022planes} and SS3D~\cite{alwala2022pre}.

\noindent \textbf{SSMP} learns single-view supervised voxel reconstruction via adversarial regularization.\footnote{SSMP has an optional test-time optimization to further convert voxels into meshes. Using the authors' public implementation of this optimization did not improve their results in our experiments, so we use their voxel prediction for evaluation.} We compare our method to SSMP over Pix3D and Pascal3D+.

\noindent \textbf{Cat3D} learns multi-class shape reconstruction without using any 3D/viewpoint annotation. It uses a similar implicit SDF representation. We compare our method to Cat3D over Pix3D and Pascal3D+. 

\noindent \textbf{SS3D} learns single-view supervised implicit shape reconstruction by pretraining on ShapeNet first. They use a per-instance camera multiplex optimization, which is too computationally expensive for us to train their model (based on their paper, even training on a single category-specific model takes 64 V100 days on average). Therefore, we use their publicly available pretrained weights instead and evaluate their method on Pascal3D+ by selecting 11 categories that SS3D has seen during training. Additionally, because SS3D cannot predict viewpoints during inference, we use the brute-force evaluation to evaluate shape reconstruction when comparing our method to SS3D. Specifically, for each instance, we align the scales of the reconstructed shape and the GT shape, and search for the rotation that leads to the lowest Chamfer Distance. We also compare our method to SS3D on OpenImages qualitatively. 

\subsection{Pix3D}
\begin{figure}[t]
\centering
	\includegraphics[width=1\linewidth]{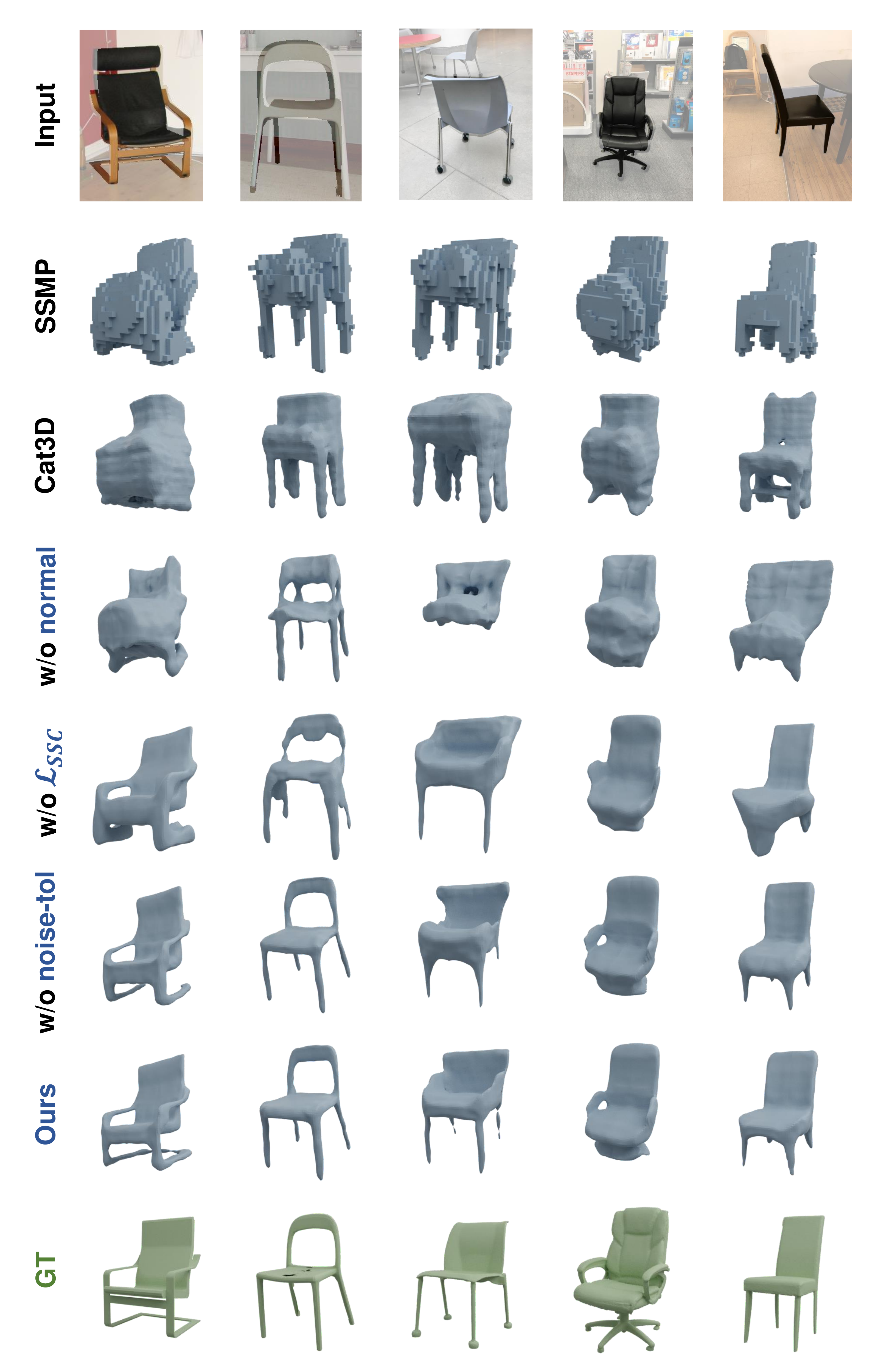}
	\caption{\textbf{Qualitative comparison on Pix3D.} Our method learns better global 3D structure and shape details than other baselines.}
	\label{fig:pix3d}
\end{figure}

We perform experiments on Pix3D and show quantitative and qualitative results in~\cref{table:pix3d} and~\cref{fig:pix3d}.

\noindent \textbf{Ablation Study.} 
We first analyze the results of ablating the techniques we propose.
In~\cref{table:pix3d}, `w/o $\mathcal{L}_{SSC}$' refers to our model without the CLIP-based semantic constraint, `w/o normal' refers to our model without the geometric constraint, 'w/o noise-tol' refers to our model without outlier dropping in the geometric constraint. Because we do not perform synthetic pretraining on Pix3D, we find the final performance of the baselines sensitive to the initialization. Therefore we average the quantitative results over 5 runs with different random seeds. By comparing the baselines to our final model, we clearly see our semantic and geometric constraints improve the reconstruction performance, and the outlier dropping benefits the shape learning as well. In the qualitative results (\cref{fig:pix3d}), we find our semantic constraint leads to better global structures, while our geometric constraint significantly improves the reconstruction of object surfaces. 

\noindent \textbf{SOTA Comparison.} Comparing with SOTA methods (\cref{table:pix3d}, last 3 rows), we see our approach outperforms Cat3D and SSMP significantly. Qualitatively, our approach captures better overall shape topology and local geometric details than Cat3D and SSMP. These results all demonstrate the effectiveness of our proposed method. Note that for comparisons on this dataset there is no synthetic pre-training for any methods.

\subsection{Pascal3D+}
\begin{figure}[t]
\centering
	\includegraphics[width=1\linewidth]{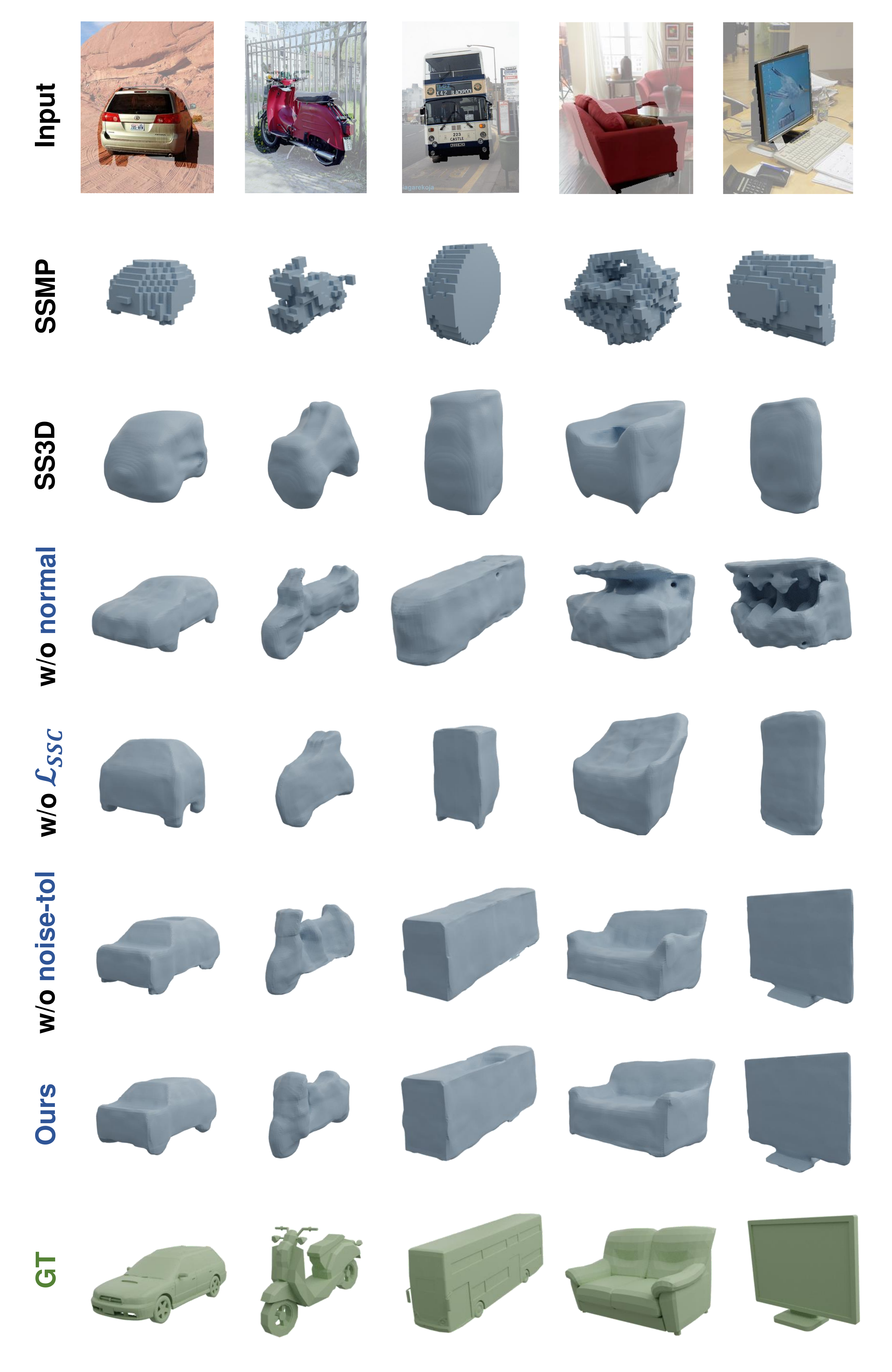}
	\caption{\textbf{Qualitative comparison on Pascal3D.} Our method learns better global 3D structure and shape details than other baselines.}
	\label{fig:pascal3d}
	\vspace{-15pt}
\end{figure}

We perform experiments on Pascal3D+ and show quantitative and qualitative results in~\cref{table:pascal3d-1},~\cref{table:pascal3d-2} and~\cref{fig:pascal3d}.
\begin{table}
\centering
\caption{\textbf{Quantitative results on Pascal3D+.} Our method performs favorably to baselines and other SOTA methods. }
\begin{tabular}{lccc|c}
\hline
Methods                     & FS@1$\uparrow$        & FS@5$\uparrow$    & FS@10$\uparrow$       & CD$\downarrow$    \\ \hline
w/o $\mathcal{L}_{SSC}$     & 0.1363                & 0.5268            & 0.7307                & 0.898             \\ 
w/o normal                  & 0.1185                & 0.4648            & 0.6712                & 0.952             \\ 
w/o noise-tol               & \textbf{0.1548}       & 0.5875            & 0.7874                & 0.707             \\ 
Ours                        & 0.1519                & \textbf{0.5914}   & \textbf{0.7981}       & \textbf{0.693}    \\ \hline
Cat3D~\cite{huang2022planes}& 0.0858                & 0.3977            & 0.6155                & 1.118             \\
SSMP~\cite{ye2021shelf}     & 0.1014                & 0.4366            & 0.6614                & 1.000             \\ \hline
\end{tabular}
\label{table:pascal3d-1}
\end{table}

\begin{table}
\centering
\caption{\textbf{Additional quantitative comparison to SS3D on Pascal3D+.} Shapes are aligned via brute-force search. Our method performs favorably to SS3D on Pascal3D+. }
\begin{tabular}{lccc|c}
\hline
Methods                     & FS@1$\uparrow$        & FS@5$\uparrow$    & FS@10$\uparrow$       & CD$\downarrow$    \\ \hline
SS3D~\cite{alwala2022pre}   & 0.0533                & 0.4879            & 0.7702                & 0.696             \\ 
Ours                        & \textbf{0.0585}       & \textbf{0.5388}   & \textbf{0.8507}       & \textbf{0.572}    \\ \hline
\end{tabular}
\label{table:pascal3d-2}
\end{table}

\noindent \textbf{Ablation Study.} 
We perform a similar ablation to Pix3D on Pascal3D+. The results corroborate the findings from Pix3D; we verify the effectiveness of our proposed techniques both quantitatively (\cref{table:pascal3d-1}) and qualitatively (\cref{fig:pascal3d}).

\noindent \textbf{SOTA Comparison.} 
We first compare our method with Cat3D and SSMP. Due to the instability of adversarial regularization and the lack of synthetic pretraining, both methods cannot scale up well to more diverse real-world scenarios and compare poorly to our method as demonstrated in~\cref{table:pascal3d-1}. In~\cref{table:pascal3d-2}, we further compare to SS3D which also uses synthetic pretraining. Because SS3D cannot predict viewpoints, we evaluate reconstructed shapes via the brute-force pose alignment for both our method and SS3D. In the quantitative comparison, our method outperforms SS3D by a large margin; we also learn better global structures and more accurate local details in the qualitative examples, as shown in~\cref{fig:pascal3d}.

\subsection{OpenImages}


We perform experiments on OpenImages and show a qualitative comparison to SS3D in~\cref{fig:openimages} across various categories. As shown in the figure, our method performs favorably to SS3D by reconstructing more accurate shapes both globally and locally. These results verify the effectiveness and scalability of our method.

\subsection{Limitations}

Although the results we achieve are promising, we find our method still can not work well for categories that are often occluded, or for largely deformable categories. 
We also do not handle the shape misalignment in our semantic constraint explicitly, which can be detrimental for categories with complex/deformable shapes. 
We think future exploration along these directions would be exciting.
\section{Conclusion}
\label{sec:conclusion}
We present a novel model that reconstructs 3D object shapes over real-world single-view images in a scalable way. 
Our model is driven by two key techniques, the CLIP-based semantic constraint and the local geometric constraint.
These two techniques significantly benefit the global shape understanding and local geometry reconstruction. 
They enable us to achieve SOTA performance on three challenging real-world datasets containing various objects. 

\noindent \textbf{Acknowledgement:} This work was supported in part by NIH R01HD104624-01A1 and a gift from Google.

{\small
\bibliographystyle{ieee_fullname}
\bibliography{refs}
}

\clearpage
\appendix
\section{Generalization performance}
\begin{figure}[h]
\centering
	\includegraphics[width=1\linewidth]{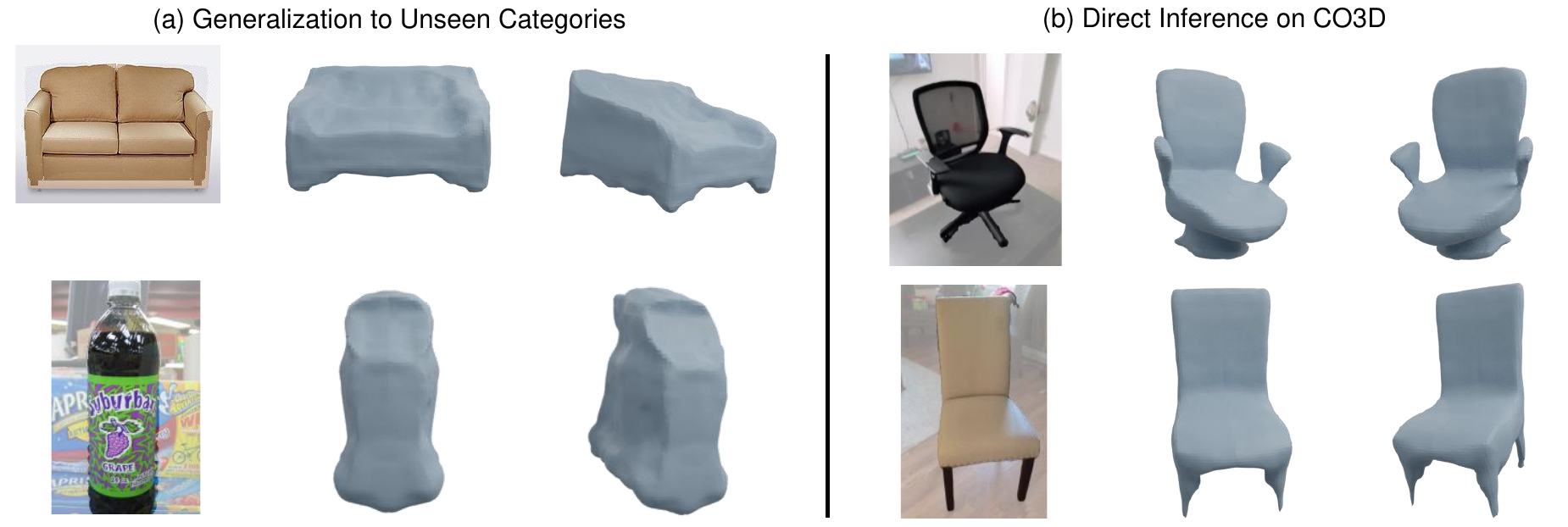}
	\caption{(a) Reconstruction on unseen categories. (b) Inference of our Pix3D-trained model on CO3D chairs.}
	\label{fig:rebuttal-quali}
\end{figure}

\noindent \textbf{Performance on unseen categories.}
We train on 6 categories (car, chair, diningtable, motorbike, train, tvmonitor) of Pascal3D+ and test on other unseen categories (see \cref{fig:rebuttal-quali} (a)). We find our model can generalize to categories that are highly related to at least one training category (e.g. sofa - average CD 0.571), and does not generalize as well to categories less related to training categories (e.g. bottle - average CD 1.020).

\begin{figure}[h]
\centering
	\includegraphics[width=1\linewidth]{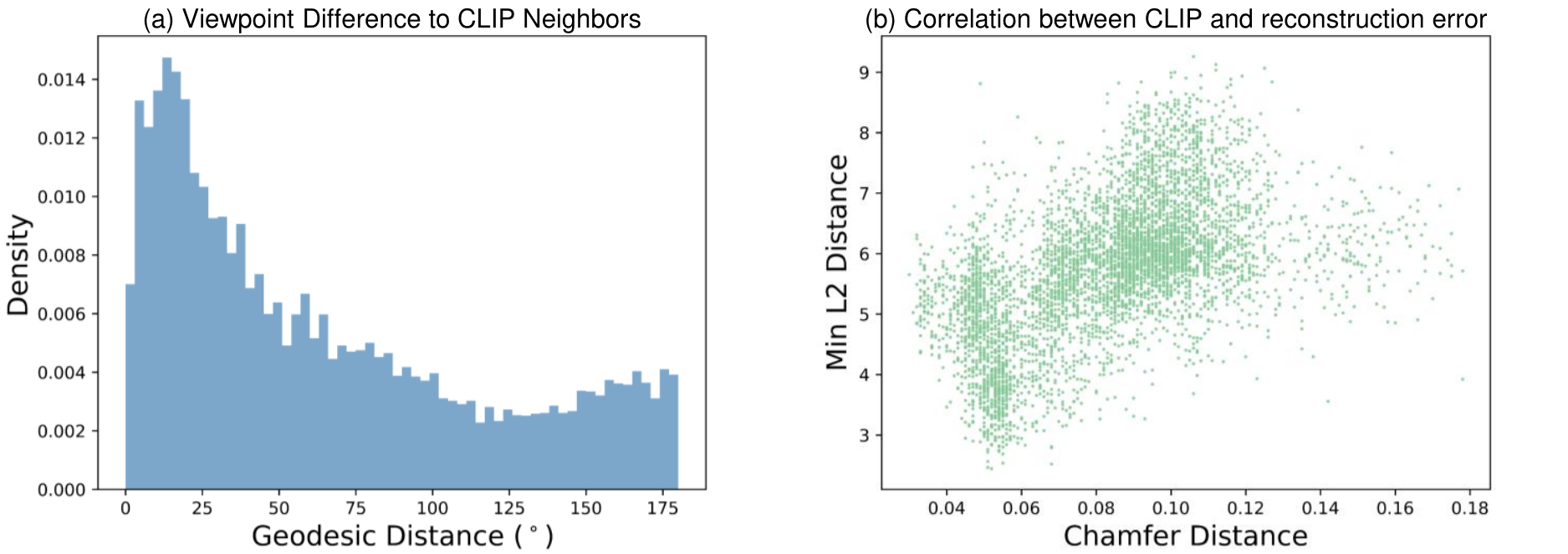}
	\caption{(a) Histogram of viewpoint distance from query images to 5 CLIP neighbors on Pix3D. (b) Correlation between reconstruction error and minimum CLIP distance on unseen categories.}
	\label{fig:rebuttal-vis}
\end{figure}

We further quantify the relationship between generalization performance and semantic relevance. We analyze the correlation between reconstruction error and minimum CLIP distance from each test sample to training images. As in \cref{fig:rebuttal-vis} (b), there exists a clear positive correlation (Pearson coefficient $\rho$ = 0.53) between the two variables. This verifies our model often generalizes better to samples that are more semantically related to the seen categories.

\noindent \textbf{Direct inference on in-the-wild data.}
Our method can also reconstruct faithful shapes under reasonable domain gaps. We test our Pix3D model directly on CO3D chair images (without fine-tuning) and find most reconstructions are reasonable. See \cref{fig:rebuttal-quali} (b) for examples.

\section{Additional Analysis of the Model}
\noindent \textbf{Viewpoint robustness of CLIP embeddings.}
We quantitatively evaluate the viewpoint robustness of CLIP embeddings on Pix3D chairs. In our experiments, CLIP can find a significant number of neighbors with distinct viewpoints (see \cref{fig:rebuttal-vis} (a)). Geodesic distance is the minimal angular difference between two rotations. The average geodesic distance from query images to CLIP neighbors is 64\degree, and 67\% of the query images have at least one neighbor with distinct pose (at least 90\degree away). 

\noindent \textbf{Robustness to corrupted masks.}
To evaluate the performance under corrupted masks, we replace the Pix3D masks with masks corrupted by perlin noise and then train/test our models under different level of pixel corruption percentages. Under \textcolor{darkred}{0}/\textcolor{darkorange}{5}/\textcolor{darkgolden}{10}/\textcolor{ForestGreen}{20}/\textcolor{NavyBlue}{30}/\textcolor{Purple}{50}\% corruption level, the chamfer distance is \textcolor{darkred}{0.612}/\textcolor{darkorange}{0.626}/\textcolor{darkgolden}{0.632}/\textcolor{ForestGreen}{0.651}/\textcolor{NavyBlue}{0.689}/\textcolor{Purple}{0.763} respectively. This experiment shows that our model is not significantly affected by mild to moderate mask inaccuracy.

\noindent \textbf{Performance with GT viewpoints.}
We further evaluate our model when GT viewpoints are given during training. An image can be explained by infinite combinations of shapes and viewpoints. When GT viewpoints are given, such entanglement is resolved and the learning will be much easier. Our model trained with GT viewpoints obtains average CD of 0.418 on Pix3D (vs. 0.612 w/o GT viewpoint).

\noindent \textbf{Additional discussion on retrieval methods.}
Comparing reconstruction methods to retrieval methods has been one of the central topics in the area of single-view shape reconstruction~\cite{tatarchenko2019single}. Based on the finding about CLIP's relationship to shape in our paper, it would be natural to consider the retrieval baseline using CLIP. While retrieving shapes with CLIP is an interesting direction, we would like to emphasize that it is not directly comparable to our proposed reconstruction method. Retrieval methods require a \textbf{large paired} image-3D shape database, similar to the non-scalable fully supervised 3D reconstruction setting. In contrast, our method only requires single-view 2D images during training, allowing it to learn to reconstruct objects from datasets like OpenImages for which there are no paired 3D shapes. Such datasets without any geometric annotations are our main application domain, and retrieval methods cannot be applied to these datasets. 

\section{Additional Implementation Details.}
\label{sec:addn-details}
\noindent \textbf{Implicit representation and rendering.}
The surface representation and texture rendering follow~\cite{yariv2021volume,yu2022monosdf}. We use an implicit SDF field and convert it to densities for volumetric rendering. The conversion from SDF to densities is done via the Cumulative Distribution Function (CDF) of the Laplace distribution:
\begin{equation}
  \sigma(s) =
    \begin{cases}
      \frac{1}{\beta} \cdot \frac{1}{2} \exp(\frac{s}{\beta}) & \text{if $s \leq 0$}\\
      \frac{1}{\beta} \cdot (1 - \frac{1}{2} \exp(-\frac{s}{\beta})) & \text{if $s > 0$}
    \end{cases},
\end{equation}
where $s$ is the SDF.
Because our focus is shape, and learning radiance is challenging without viewpoint annotation, we represent texture as an RGB field without any view-dependency. The surface normal rendering follows MonoSDF~\cite{yu2022monosdf}, where the local normal vectors are estimated by the gradient of the SDF field and aggregated via the standard volume rendering. 

\noindent \textbf{Uniform viewpoint prior.}
We use a uniform prior to regularize the viewpoint learning, which helps to prevent the rotation estimation from degeneration. Specifically, for each minibatch during training, we estimate the empirical distribution of the predicted azimuth. We then minimize the Earth-Mover Distance (EMD) between the empirical azimuth distribution and a uniform prior within $[0\degree, 360\degree]$.

\section{Additional Results on OpenImages}
\label{sec:openimage-extra}
We show more reconstruction results of our model trained on all 53 OpenImages~\cite{kuznetsova2020open} categories used in~\cite{ye2021shelf}. We further compare with SS3D~\cite{alwala2022pre} qualitatively, where our model demonstrates state-of-the-art reconstruction performance. Note the training is category-specific and the performance may be further improved by training a joint model via distillation~\cite{alwala2022pre}, which is parallel to our research direction here.

\begin{figure*}[h]
\centering
	\includegraphics[width=1\linewidth]{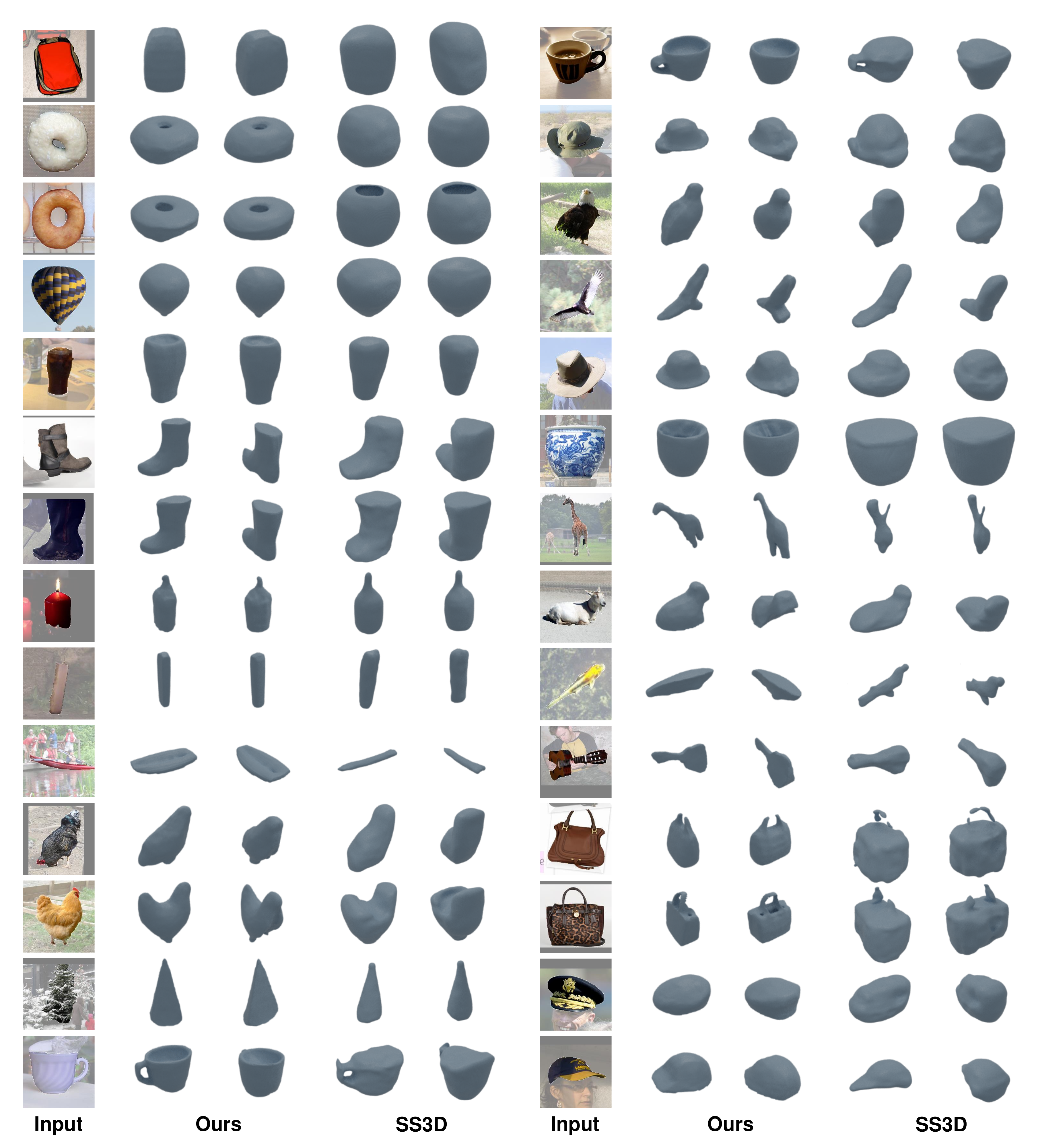}
	\caption{Additional qualitative results and comparison on full OpenImages.}
	\label{fig:openimages-full-1}
	\vspace{-16pt}
\end{figure*}

\begin{figure*}[h]
\centering
	\includegraphics[width=1\linewidth]{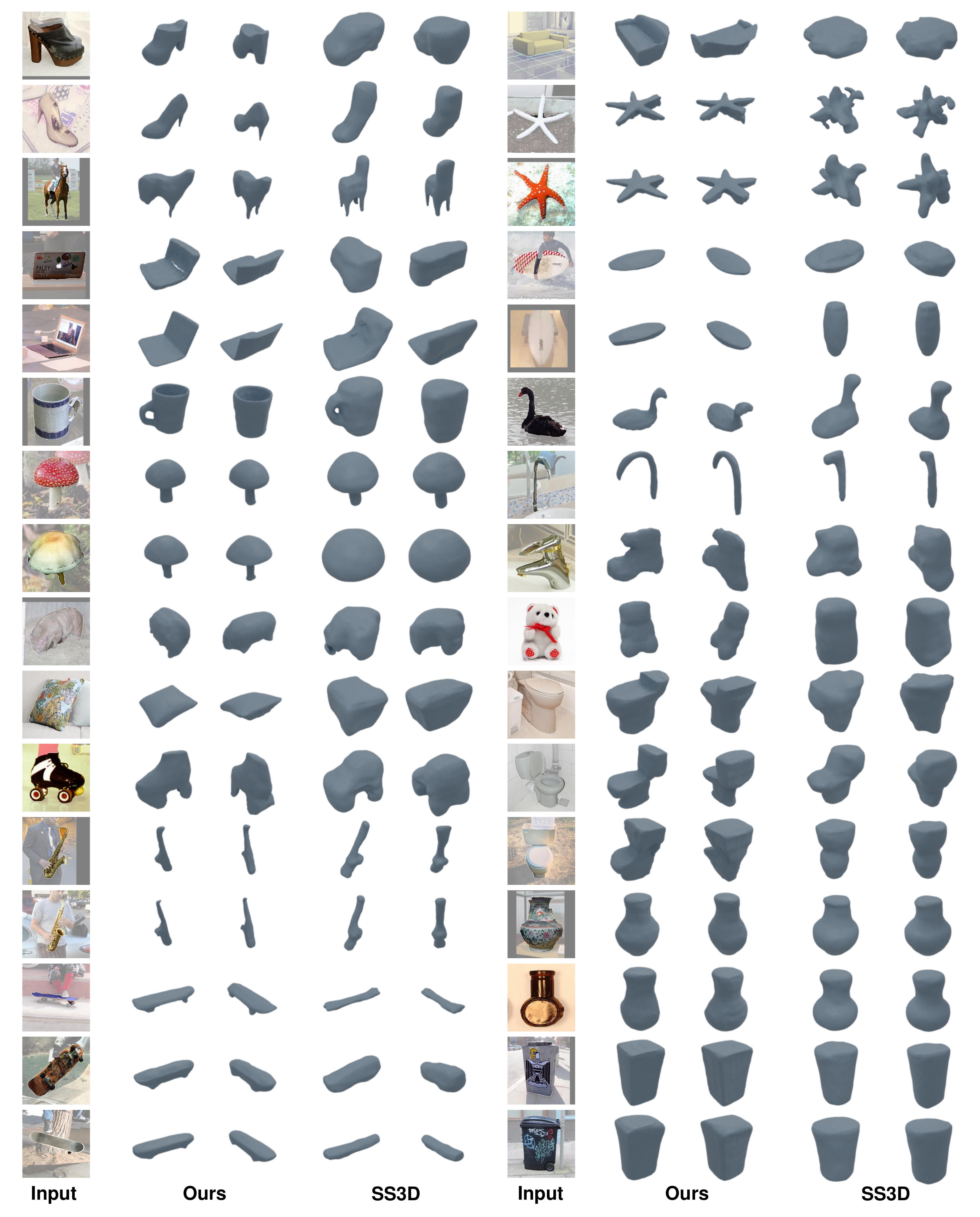}
	\caption{Additional qualitative results and comparison on full OpenImages.}
	\label{fig:openimages-full-2}
	\vspace{-16pt}
\end{figure*}

\end{document}